%% file: main.tex
\documentclass{article}
\pdfoutput=1 
\usepackage{ijcai19}
\usepackage{mathtools}
\usepackage{mathrsfs}

\usepackage{times}
\usepackage{soul}
\usepackage{url}
\usepackage[hidelinks]{hyperref}
\usepackage[utf8]{inputenc}
\usepackage[small]{caption}
\usepackage{graphicx}
\usepackage{amssymb}
\usepackage{amsmath}
\usepackage{booktabs}
\usepackage{algorithm}
\usepackage[algo2e,linesnumbered,lined,ruled,vlined]{algorithm2e}
\usepackage[noend]{algpseudocode}
\usepackage{verbatim}
\usepackage{arydshln}

\title{Efficient  Predicate Invention using Shared NeMuS}

\author{
Edjard Mota$^1$
\and
Jacob M. Howe$^2$
\and
Ana Schramm$^1$
\and
Artur d'Avila Garcez$^2$\\
\affiliations
$^1$Institute of Computing, Federal University of Amazonas, Manaus - Brazil\\
$^2$Department of Computer Science, City, University of London, UK \\
\emails
\{edjard, acms\}@icomp.ufam.edu.br, \{j.m.howe, A.GARCEZ\}@city.ac.uk
}
\newcommand{\vect}[1]{\boldsymbol{#1}}

\begin{document}

\maketitle

   \input{./abstract.tex}
   \input{./introduction.tex}
   \input{./background.tex}

   \input{./inductive.tex}

   \input{./predicateInvention.tex}
   \input{./related.tex}

   \input{./conclusions.tex}

\bibliographystyle{named}
\bibliography{nesy_19.bib}

\end{document}

%% file: abstract.tex
\begin{abstract}
Amao is a cognitive agent framework that tackles the invention of predicates with a different strat- egy as compared to recent advances in Inductive Logic Programming (ILP) approaches like Meta- Intepretive Learning (MIL) technique. It uses a Neural Multi-Space (NeMuS) graph structure to anti-unify atoms from the Herbrand base, which passes in the inductive momentum check. Induc- tive Clause Learning (ICL), as it is called, is ex- tended here by using the weights of logical compo- nents, already present in NeMuS, to support induc- tive learning by expanding clause candidates with anti-unified atoms. An efficient invention mecha- nism is achieved, including the learning of recur- sive hypotheses, while restricting the shape of the hypothesis by adding bias definitions or idiosyn- crasies of the language.

\end{abstract}

%% file: introduction.tex
\section{Introduction}

One of the key challenges in Inductive Logic Programming (ILP) is finding good heuristics 
to search the hypothesis space. In Standard ILP, a good heuristic is one that can arrive quickly at a hypothesis that is both successful and succinct. To achieve this, the efficiency of hypothesis generation depends on the partial or even total order over the Herbrand Base to constrain deduction operations. This work presents a new approach called  Inductive Clause Learning (ICL) building on \cite{nemus4NeSy16}, which 
introduces a data structure named Neural Multi-Space (NeMuS) used in Amao, a neural-symbolic reasoning platform that performs symbolic reasoning  on structured 
clauses via Linear Resolution \cite{ar:resolution}.

 Inspired by 
\cite{boyer-moore}, NeMuS is a shared multi-space representation for a portion of 
first-order logic designed for use with machine learning and neural network 
methods. Such a structure contains weightings on individual elements (atoms, 
predicates, functions and constants or variables) to help guide the use of these elements in tasks such as theorem proving, as well as in using them to guide the search in the hypothesis space and improve the efficiency and success of inductive learning. Although it has some similarities to ILP, the hypotheses search mechanism is fundamentally different. It uses the Herbrand 
Base (HB) to build up hypothesis candidates using inverse unification
(adapted from \cite{idestam93generalization}), and prunes away meaningless hypotheses as a result of {\em inductive momentum\/} between predicates connected to positive and negative examples. In \cite{ilSNeMuS4NeSy17} inverse or anti-unification
\cite{idestam93generalization} was added to allow induction of general  
rules from ground clauses, which is supported by the idea of {\em regions of concepts\/}. However, the inductive learning algorithm presented did not consider an adequate representation and use of \emph{bias}, and the invention of predicates called \emph{predicate invention}. Here we show how this can be achieved without using meta-interpreter level
of bias specification or reasoning. It is important to note that weights are still not automatically used, but are taken into account when apparently unconnected literals have a common predicate name.

This paper makes the following contributions: it demonstrates that it is possible to have predicate invention without the use of meta-rules, and consequently, of Meta-Interpretive Learning. It shows that the NeMuS structure can be used for this purpose without generating numerous meaningless hypotheses, as the invention is made during Inductive Clause Learning. For that, we use bias or automated predicate generation. Finally, it demonstrates how invention of recursive rules takes advantage of weights of the logical component representation within NeMuS.

The remainder of this paper is structured as follows: section~\ref{background} gives some
brief background on inductive logic programming and the Shared NeMuS data structure, 
sections~\ref{inductive} and \ref{sec:ICL-invent} describe the implementation of inductive 
learning in Amao using the Shared NeMuS data structure, then section~\ref{related}
describes some related work and section~\ref{conclusions} discusses the work presented.

%% file: background.tex
\section{Background}\label{background}


\subsection{Inductive Logic Programming (ILP)} 

The Inductive Logic Programming (ILP) main challenge, as defined in
 \cite{muggleton91inductive}, is to search for a logical description (a hypothesis, 
 $H$) of a target concept, based on set of (positive and negative) examples
along with a set called background  knowledge ($BK$). The central idea is that 
$H$ is a \emph{consistent hypothesis}, i.e. $BK$ plus the hypothesis $H$ entails 
the positive examples ($e^{+}$), whilst does not entail the negative ones ($e^{-}$).  
Formally, $BK \cup \{H \}\vdash e^{+}$ and $BK \cup \{H \} \not \vdash e^{-}$. 

Typical ILP systems implement search strategies over the space of all possible 
hypotheses. To reduce search complexity, such mechanisms rely on partial order of 
$\theta$-subsumption \cite{nienhuys1997foundations}, or on a 
total ordering over the Herbrand Base to constrain deductive, abductive and inductive 
operations \cite{muggleton15meta}. As a side-effect, the space of hypotheses 
grows even more due to quantification over meta-rules by the meta-interpretive
process.

The  Inductive Clause Learning (ICL) technique is fundamentally different, while 
its learning results are similar to ILP and MIL. It does not generate hypotheses 
to then test whether $BK \cup H$ entails positive examples but not the negative ones.
Instead, ICL anticipates the elimination of inconsistent hypotheses at each 
induction step by \textit{colliding} atoms obtained from a search across bindings 
of constants from $e^{+}$ and $e^{-}$. This is possible because NeMuS is a network 
of shared spaces (for constant terms, predicates and clauses), interconnected through 
weighted bindings pointing to the target space in which an occurrence of an element 
appears. 
In what follows this is briefly described.

\subsection{Shared NeMuS}
\label{subsec:revisitingNeMuS}

NeMuS is an ordered space for components of a first-order language: variables (space 0), 
atomic constants of the Herbrand Universe (space 1), functions (space 2, suppressed here), 
predicates with literal instances (space 3), and clauses (space 4), and so on. In what follows 
vectors are written $\vect{v}$, and $\vect{v}[i]$ or $\vect{v}_i$ is used to refer to an 
element of a vector at position $i$.

Each logical element is described by a vector called T-Node, and in particular each element 
is uniquely identified by an integer code (an index) within its space.  In addition, a T-Node 
identifies the lexicographic occurrence of the element, and (when appropriate) an attribute position. 

\newtheorem{TNode}{Definition}
\begin{TNode}[T-Node]
Let $c \in \mathbb{Z}$, $a, i \in \mathbb{Z}^{+}$ and  $h, \in \mathbb{N}$. A {\em T-Node (target node)\/}  is a 
quadruple $(h, c, i, a)$ that identifies an object at space $h$, with code $c$ and occurrence 
$i$, at attribute position $a$ (when it applies, otherwise 1). ${\cal T}_{N}$ is the set of all T-Nodes.
For a vector  $\vect{x}$, of T-Nodes, with size $n$, and $c$ is a code
occurring in an element of  $\vect{x}$, then $\iota(c, \vect{x}) = k$ is the index of $c$ within 
the T-Node, $0 \leq k \leq (n-1)$.
\end{TNode}

As T-Nodes are the building block of our approach,  all other elements follows from it and we
describe as follows.

\begin{description}
\item [NeMuS Binding] is an indexed pair $(p,w)_k$ in which 
        $p \in {\cal T}_{N}$, $w \in \mathbb{R}$ and $k \in \mathbb{Z}^{+}$, such that
        $n_{h}(p) = h$,  $n_{c}(p) = c$, $n_{a}(p) = a$ and $n_i(p) = i$. It represents the importance 
        $w$ of object $k$ over occurrence $n_i(p)$ of object $n_{c}(p)$ at space $n_{h}(p)$ in 
        position $n_{a}(p)$. 
\end{description}

\begin{description}
\item [Variable Space (0)] is a vector  $ \vect{V} = [ \vect{y}_1, \ldots,  \vect{y}_n ]$, in which each 
        $\vect{y}_i$ is a vector of bindings. The elements of the variable space represent all of the 
        occurrences of a variables. The logical scope of a variable  $X$ is identified
        by the instances of its bindings.
\end{description}
\begin{description}
\item [Constant Space (1)] is a vector $\vect{C} = [ \vect{x}_1, \ldots,  \vect{x}_m ]$ , in which
         every $\vect{x}_i$ is a vector of bindings. The function $\beta$ maps a constant $i$ to 
         the vector of  its bindings $\vec{x_i}$, as above.
\end{description}

Compounds  (functions, predicates and clauses), are in higher spaces.
Their logical components are formed by a vector of T-Nodes (one for each argument), and 
a vector of NeMuS bindings (simply bindings) to represent their instances.

\begin{description}
\item [Compound] in NeMuS is a vector of T-Nodes, i.e. $\vect{x}^{i}_a = [c_1, \ldots, c_m]$, 
         so that each $c_j \in {\cal T}_{N}$, and it represents an attribute  of a compound logical
         expression coded as $i$. 
\item [Instance Space] (I-Space) of a compound $i$ is the pair $(\vect{x}^{i}_{a}, \vect{w}_i)$ 
         in which $ \vect{w}_i$ is a  vector of bindings.  A vector of I-Spaces is a NeMuS    
         \emph{Compound Space (C-Space)}.
\end{description}

A literal (predicate instance), is an element of an I-Space, and so
the predicate space is simply a C-Space. Seen as compounds, clauses' attributes are the 
literals composing such clauses. 

\begin{description}
\item [Predicate Space (3)] is a pair $(\vect{C}^{+}_p, \vect{C}^{-}_p)$ in which $\vect{C}^{+}_p$ 
         and $\vect{C}^{-}_p$ are vectors of C-spaces.  
\item [Clause Space (4)]  is a vector of C-spaces such that every pair in the vector shall be  
         $(\vect{x}^{i}_{a},[ ])$. 
\end{description}

Note that the order of each space is only defined when they are gathered in the following
structure.

\newtheorem{sharedNeMuS}[TNode]{Definition}
\begin{sharedNeMuS}[Shared NeMuS]
A \emph{Shared NeMuS} for a set of coded first-order expressions is a 4-tuple (assuming no functions),
${\cal N}: \langle \mathcal{V}, \mathcal{S}, \mathcal{P}, \mathcal{C} \rangle$, in which $ \mathcal{V}$ is the variable space, $\mathcal{S}$ is the constant space, 
$\mathcal{P}$ is the predicate space and $\mathcal{C}$ is the clause space. 
\end{sharedNeMuS}

The next section describes how inductive learning is performed in Amao using NeMuS structure.

%% file: inductive.tex
\section{NeMuS-based Inductive Learning}\label{inductive}

ICL is based on the concept of Least Herbrand Model 
(LHM) \cite{lloyd93foundations} to anticipate the elimination of inconsistent hypotheses, 
at each induction step, before they are fully generated. 
This is done by \textit{colliding}, via computing the \emph{inductive momentum}, atoms 
obtained from bindings of arguments from $e^{+}$ (candidates to compose LHM) and $e^{-}$. 
Then, a pattern of linkage across verified literals is identified, anti-unification  
(adapted from \cite{idestam93generalization}) is applied, and a conjecture is generated. 
The process repeats until the conjecture becomes a closed and consistent hypothesis. 


\subsection{Inductive Learning from the Herbrand Base}
\label{aspectsILHB}

\begin{figure}[h!]
    \centering
    \includegraphics[scale=0.9]{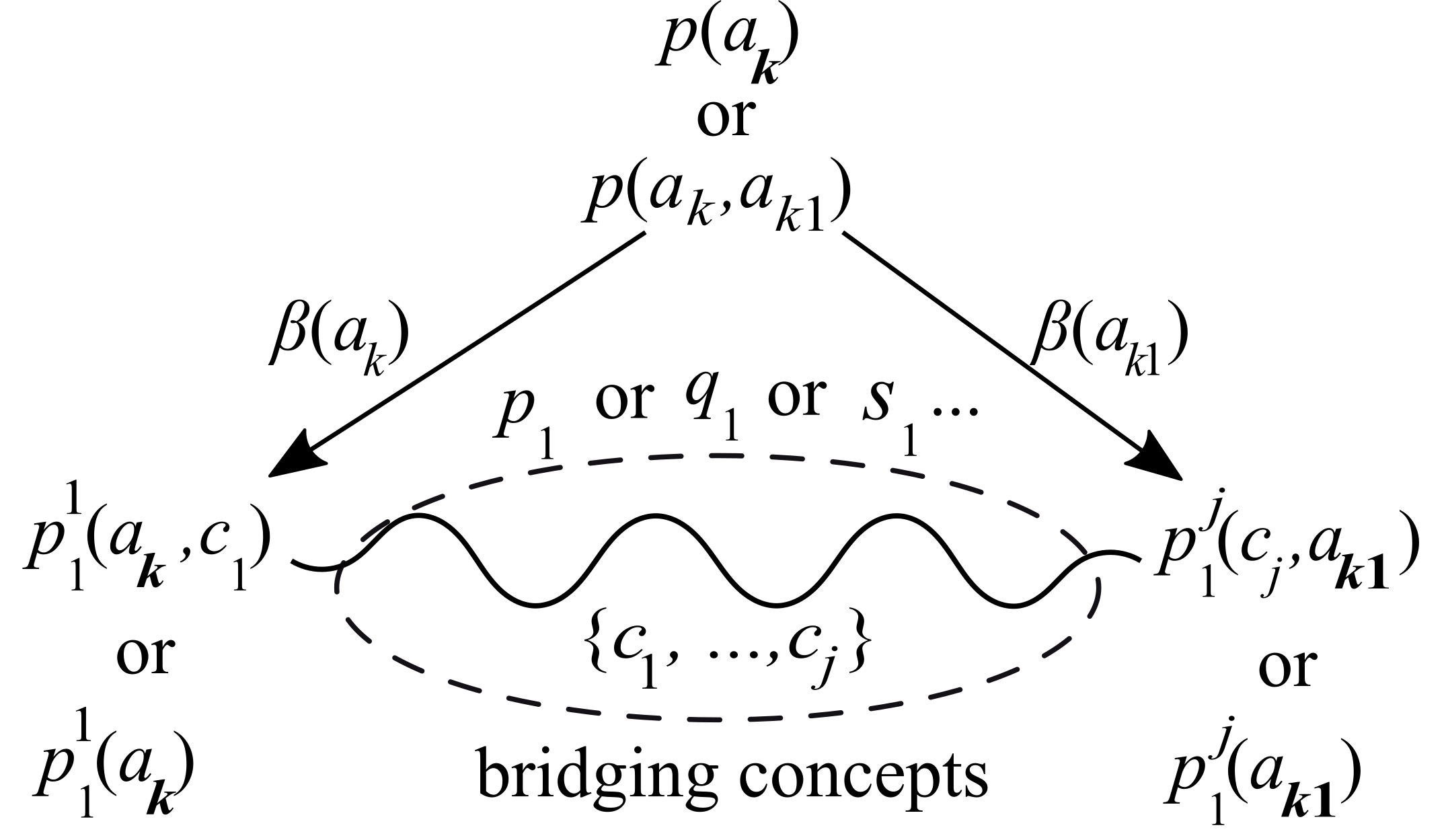}
    \caption{Portion of the HB with bindings of $t$'s attributes.}
    \label{fig:AllHookPathsFromExamples}
\end{figure}

The problem of inductive learning involves a knowledge base of predicates, 
called background knowledge ($BK$), a set $E$ of examples that the logical 
description $H$ of the target concept ($t$) should prove (positive examples, $e^{+}$) 
and a set of examples that the target concept should not prove (negative examples, 
$e^{-}$). Figure~\ref{fig:AllHookPathsFromExamples} depicts a possible
Herbrand Base that may explain  how $H$ entails a positive example for the 
concept $t$. In its turn, $t$ can be unary ($p(a_k)$), binary ($p(a_k,a_{k1})$),
etc.

From attribute bindings of $t$, i.e. $\beta(a_k)$ and possibly $\beta(a_{k1})$,
$t$ is connected with attribute mates' bindings. Such connections bridge
all concepts they may appear in ($p_1$, $q_1$, etc.), like a 
path $\{c_1, \ldots, c_j\}$ in a graph, until it reaches the binding concept of 
$t$'s last attribute. The interconnected concepts form a  \emph{linkage pattern}.
For example, when $p_1$, $q_1$ etc., are all the same concept,  then a recursive 
hypothesis may be generated.
Invention and hypotheses generation and will always take place in the \textit{bridging concepts}' region 
(including the initial and last binding concepts).

The induction method is based on the following aspects: (a)  \emph{Inductive 
momentum} that iteratively ``selects" only those atoms not likely to entail $e^{-}$,
(b) \emph{linkage patterns} among atoms passed (a), based on internal connections 
(bridging  concepts); (c) \emph{anti-unification} substitutes constants in an atom 
from the Herbrand Base by variables (and optionally (d)   \emph{category-first ordering},
\cite{categoryBaseLNeMuS4NeSy17}, useful when $BK$ contains monadic 
definitions of categories). 

\subsection{Linkage Patterns and Hypothesis}
\label{sec:linear-linkpattern}


A special form of intersection $\rho$  identifies the common terms between both literals.
For instance, in Figure~\ref{fig:AllHookPathsFromExamples},  suppose $j=1$, i.e. 
bridging concepts has just two literals of the same concept $p_1$:  $p_1(a_k, c_1)$  and 
$p_1(c_{1}, a_{k1})$.  Then $\rho(p^{1}_{1},p^{2}_{1})=c_1$.

\newtheorem{hook4Indution}[TNode]{Definition}
\begin{hook4Indution}[Linkage and Hook-Terms]
Let $p$ and $q$ be two predicates of a $BK$. There 
is a \emph{Linkage} between $p$ and $q$ if a same constant, $t_h$, appears (at least) 
once in ground instances of $p$ and $q$. We call $t_h$ a \emph{hook-term}, computed
by $t_h = \rho(p,q)$. The \emph{attribute mates}  $t_h$ w.r.t. an atom $p$,
written $\overline{\rho(p,q)}_p$ is a set of terms occurring in $p$, but not in $q$.
\end{hook4Indution}

With $p^{1}_{1},p^{2}_{1}$ as above, 
$c_1$ is their hook term
and $\overline{\rho(p^{1}_{1},p^{2}_{1})}_{p^{1}_{1}} = \{a_{k}\}$ and 
$\overline{\rho(p^{1}_{1},p^{2}_{1})}_{p^{2}_{1}} = \{a_{k1}\}$. This would form
the definite clause $p(a_k, a_{k1}) \leftarrow p_1(a_k, c_1) \land p_1(c_{1}, a_{k1})$,
which is not generated but it is built using anti-unification to generalize over its hooked 
ground literals.
 In the following definition we use the standard notion of a substitution 
$\theta$ as a set  of pairs of variables and terms like $\{X_1/t_1, \ldots, X_n/t_n\}$.

\newtheorem{antiUnification}[TNode]{Definition}
\begin{antiUnification}[Anti-substitution  and Anti-unification]
Let $G$ be a first-order expression with no constant term and 
$X_1, \ldots, X_n$ be free variables of $G$, $e$ is a ground first-order expression,
and $t_1, \ldots, t_n$ are constants terms of $e$.  
 An \emph{anti-substitution} is a set $\theta^{-1} = \{t_1/X_1, \ldots, t_n/X_n\}$ such that 
$G = \theta^{-1} e$, and $G$ is called a simple \emph{anti-unification} 
of $e$. The \emph{anti-unification} $f^{-1}_{\theta}$ maps a ground atom $e$ to its
corresponding \emph{anti-substitution} set that generalizes $e$, i.e. $f^{-1}_{\theta}(e) = 
\theta^{-1}$
\end{antiUnification}

Note that in the original definition of anti-unification,  \cite{idestam93generalization},
$G$ should be the generalisation of two ground expressions. Here, as we build
a hypothesis by adding literals from a definite clause the definition is

\newtheorem{ground2General}[TNode]{Definition}
\begin{ground2General}[Anti-unification on linkage terms]
Given two literals $p(a_k, a_z)$  and $q(a_z, a_{k1})$. 
A \emph{linkage term}, say $Z_0$, for their hook term $a_z$, is a variable that can be placed,
by anti-unification, in the hook's position wherever it appears in the ground instance that will 
produce two non-ground literals.
\end{ground2General}

The definite clause $p(a_k, a_{k1}) \leftarrow p_1(a_k, c_1) \land p_1(c_{1}, a_{k1})$,
from Figure~\ref{fig:AllHookPathsFromExamples} ($j=1$),  
$\theta^{-1} = \{a_k/X, a_{k1}/Y, c_{1}/Z \}$ is incrementally
anti-unified, and so the following general clause is found:
$p(X, Y) \leftarrow p_1(X, Z) \land p_1(Z, Y)$. 

This concept is the fundamental operation to generate a hypothesis because 
it generalizes ground formulas into universally quantified ones. Before describing
how negative examples are used  we have the following definition.

\newtheorem{hypothesis}[TNode]{Definition}
\begin{hypothesis}[Hypothesis]
Let $H$ be a formula with no constant term, $S$ be a set of ground atoms 
formed by concepts and constants from a Herbrand universe $H_u$, $t$ is a ground atom 
with $k$ terms (which belongs to the base of constants, $H_0$) such that $t \not \in S$. 
We say that $H$ is a \emph{hypothesis} for $t$ with respect to $S$ if and only if there 
is a set of atoms $E = \{e_1, \ldots, e_n \}$ and a $ \theta^{-1}$ such that for

\begin{enumerate}
\item every $a$ of $t$, there is some $e_i \in E$ and  $\rho(t, e_i) = a$
\item every $e_i$ and $e_{i+1}$  $\rho(e_i, e_{i+1})$ is not empty.
\item when 1 and 2 hold, then $H = \theta^{-1} (\{t\} \cup E)$.
\end{enumerate}
An \emph{open hypothesis} is one that at least one term of $t$ have not been
anti-unified. Thus, 1, 2 and 3 will always generate a closed least hypothesis.
\end{hypothesis}

Recall that \emph{learning should 
involve the generation of a hypothesis and to test it against positive and negative examples}. 
Such a ``test" could be done while the hypothesis is being generated. 
This is the fundamental role of the following concept.
Note that, in the interest of saving space, the following sections shall use logic notation. It is important, however, to recall the definitions given in section \ref{subsec:revisitingNeMuS}, as the method described runs on top of the Shared NeMuS structure.

\subsection{Inductive Momentum} 
\label{subsec:im}

In the following definition, $a_k$ is a constant originating from the path of a
positive example, and $b_k$ is another constant originating from a negative example.
 
 \newtheorem{inductiveMomentum}[TNode]{Definition}
\begin{inductiveMomentum}[Inductive Momentum]
Let $(\vect{x}^{i}_a, \vect{w}_i)$ and $(\vect{x}^{j}_a, \vect{w}_j)$ be two I-spaces
of atoms $i$ and $j$, representing $l^{+} $ and $l^{-}$ atomic formulas 
(literals) in the Herbrand base. If $\exists k$ and $m$, from $e^{+}$ and $e^{-}$,
such that $l^{+} \in \beta(k)$  and $l^{-} \in \beta(m)$,
i.e. $k$ is an element of $\vect{x}^{i}_a$ and $m$ is an element of $\vect{x}^{j}_a$, then
the \emph{inductive momentum} between $l^{+}$  and $l^{-}$ with respect to $k$ and $m$ is

\[ I_{\mu}(\vect{x}^{i}_a, \vect{x}^{j}_a)^{k}_{m} =
  \begin{cases}
    \text{inconsistent}    & \text{if } i = j \text{, and } \\ & \quad \iota(k,\vect{x}^{i}_a)  = \iota(m,\vect{x}^{j}_a)\\
    \text{consistent}  & \text{otherwise}
  \end{cases}
\]
\end{inductiveMomentum}

Note that if $i = j$, then it is assumed $ \| \vect{x}^{i}_a \|  =  \| \vect{x}^{j}_a \| $ since they 
are the same code in the predicate space. When it is clear in the context we shall simply write
$I_{\mu}(l^{+}, l^{-})^{k}_{m}$ rather than the T-Node vector notation.

\noindent
{\bf Example 1}. $BK$ is formed by ground instances of binary and 
monadic predicates (not limited to them), atoms and Herbrand universe $H_u$ as follows.

\begin{enumerate}
\item $\{p_1(a,a_1), \ldots, p_k(a_k,a)\}$, 
\item $\{q_1(a_1, b_1), \ldots, q_j(b_j,a_1)\}$, 
\item $\{r_1(c_1, a_k) \ldots, r_m(a_k,c_m)\}$, 
\item $\{t_1(b_j), \ldots, s_1(c_1),\ldots, v_1(c_m),\ldots\}$,
\item target  $p(X)$, with $e^{+}$: $p(a)$ and $e^{-}$: $\sim p(b)$. 
\end{enumerate}

When the BK is compiled its correspondent NeMuS structure is also built. 
The induction mechanism, at each step, adds to the premise of a hypothesis the 
next available atom from the bindings of a constant only if such 
an atom ``resists" the inductive momentum. 

\begin{table}[th]
\centering
\begin{tabular}{crl}
\toprule
Step & partial hypothesis    & $I_{\mu}$            \\
\midrule
1 & $p(X) \leftarrow p_1(X,Y)$             & n/a                       \\
2 & $p(X) \leftarrow p_1(X,Y) $  & $I_{\mu}(q_1(a_1,b_1), r_1(b_1,c_1))^{a_1}_{b_1}$ \\
   &       $\land q_1(Z, Y)$                    &  =  {\it consistent} \\
3 & $p(X) \leftarrow p_1(X,Y)$   & $I_{\mu}(q_1(a_1,b_2), r_2(b_2,c_2))^{a_1}_{b_1}$  \\
   &       $\land q_2(Z, Y)$                        &  =   {\it consistent} \\
$\ldots$ & $\ldots$  &  \\
$n$ & $p(X) \leftarrow p_1(X,Y) $ & $I_{\mu}(q_j(b_j, a_1), q_j(b_j,b_1))^{a_1}_{b_1}$ \\
   &      $\land q_j(Z, Y)$                           &  =  {\it inconsistent} \\
\bottomrule
\end{tabular}
\end{table}

For a partial hypothesis would be $p(X) \leftarrow p_1(X,Y) \land q_j(Z, Y)$
with $\theta^{-1} = \{ a/X, a_1/Y, b_j/Z\}$, but the equivalent path from negative example 
would reach $q_j(b_j, b_1)$. This would allow $p(b)$ to be also deduced, which is not
what it is expected from a sound hypothesis. Thus, this hypothesis is dropped. 
For this example, a
sound hypothesis could be 
$p(X) \leftarrow p_k(Y, X) \land r_1(Z, Y) \land s_1(Z)$. 


Had the target concept be $s(X)$ and positive example $s(c_1)$, then a possible 
hypothesis generated would be


$s(X) \leftarrow  s_1(X) \land r_1(X, Z_0) \land p_k(Z_0, Z_1)$. 

\subsection{Predicate Invention}
\label{sec:predicateInvent}

 Predicate invention, according to ILP definition, is a \textit{bias} defined by  the 
 user via a declarative language. It is a way to deal with predicates missing from the $BK$
 for the lack of information. Suppose that target concept $p$ of Figure~\ref{fig:AllHookPathsFromExamples}
 is $ancestor(X,Y)$, BK is the set $B = \{ father(jake, alice),  
 mother(matilda, alice), \ldots \}$.  There are two different concept relations in which the constant 
 $alice$  participates as a second attribute.There can be many instances of  both 
 $father$ and $mother$, and no constant appear as first argument of both. 
 There seems to be new concept that  \emph{captures the property that all 
 persons share when appearing as the  first attribute of either relation}. 
 
\begin{figure}[h!]
    \centering
    \includegraphics[scale=0.75]{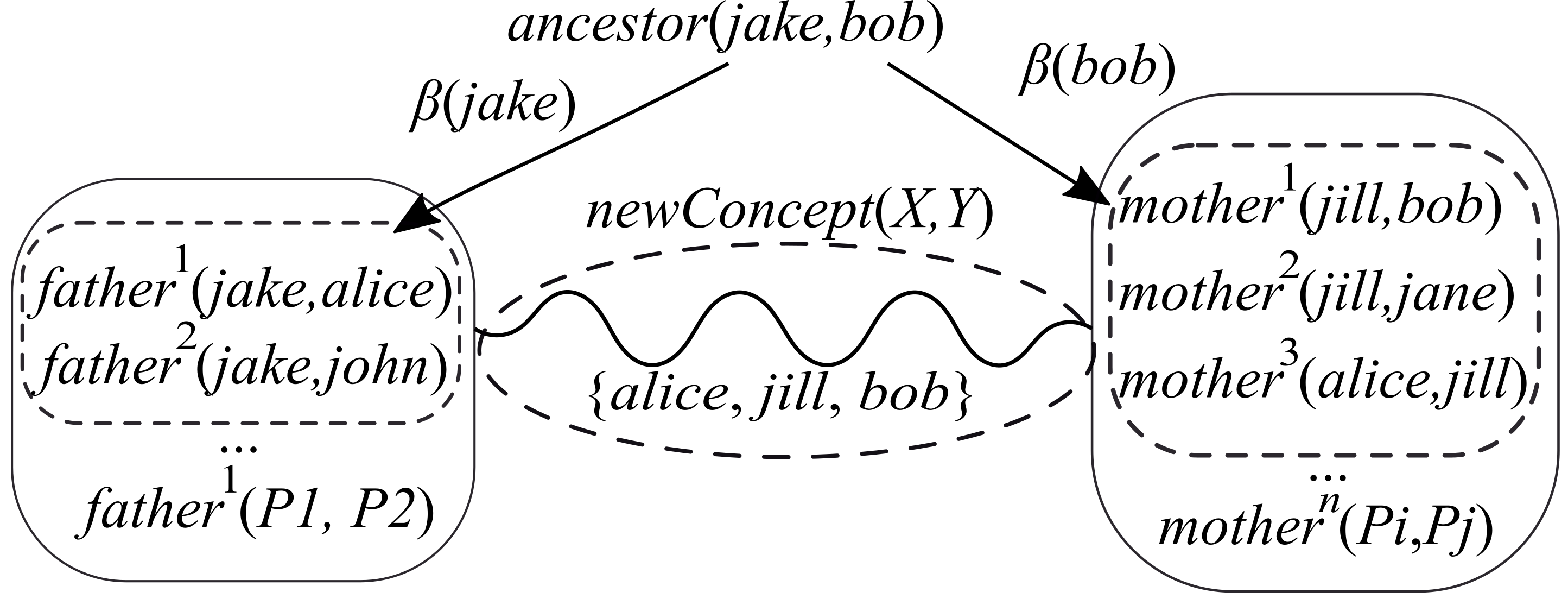}
    \caption{The invented predicate bridges the two regions of concepts.}
    \label{fig:recursivePattern}
\end{figure}

On a closer look at
Figure~\ref{fig:recursivePattern}, it is possible that our general approach to predicate 
invention generates hypotheses that do not look like what we expect. For example $ancestor(X,Y)$ 
given $ancestor(jake, bob)$, might generate $ancestor(X,Y) \leftarrow father(X,Z_0) 
\land p_1(Z_0,Y)$.

Is is assumed that two concepts, say $c_1$ and $c_2$, are ``specialisations"
of  another concept $c$ whenever there are objects appearing as second argument of both, 
but can only appear as first in one of them. This is informed to Amao as follows
 
 \vspace{11pt}
 \noindent
 {\tt Consider induction on T knowing E assuming P1 or P2 defines NewP}, 
 
 \noindent to mean 
 that the bias we are looking for is  {\tt NewP $\rightarrow$ P1} or {\tt NewP $\rightarrow$ P2}.

 \vspace{11pt}
  
We say that an \emph{invented predicate bridges two regions of concepts}, and so allowing 
a more simple generalisation of ground rules into hypothesis. This is illustrated in
Figure~\ref{fig:recursivePattern}. 

\noindent
$target(X, Y) \leftarrow newConcept(X, Z) \land target(Z, Y)$
 
Every time either or both concepts are involved in a hypothesis generation, the new concept is 
used to intentionally define the target predicate. So, from the figure above the rule base would be 

\begin{center}
$newConcept(X, Y) \leftarrow (father(X, Y) \lor mother(X, Y))$
\end{center}

Of course the new concept is parent and it is not a target concept to consider induction, 
but shall be used as a bridge or as a base form of a hypothesis, while the target shall be a 
linear or recursive linkage pattern (in our approach), or tail recursive.

%% file: predicateInvention.tex
\section{Inductive Clause Learning with Invention}
\label{sec:ICL-invent}

The method we are going to present in this section joins all ideas described in
section~\ref{inductive}.  We shall use standard logic program notation for clauses 
just for readability sake,  but recall that Amao language treats $q \leftarrow p$ as
 $q \lor \lnot p$.
The general idea of ICL can be 
summarised in three mains steps. 

\begin{enumerate}
\item to walk across the linkages found in the Herbrand Base in order to select 
         atoms as candidates for composing hypotheses, as well as those to oppose 
         the compositions
\item to compute $I_{\mu}$ of atoms as candidates for anti-unification that were selected
         from positive and negative linkages.
\item to generalize, via anti-unification, only atomic formulas likely to build consistent 
         hypotheses, i.e. those composed by atoms consistent with respect to  $I_{\mu}$         
\end{enumerate}

In the following description we shall consider a dyadic theory with no function terms/

\subsection{Selecting Candidates to Compose Hypothesis}
\label{sec:hypoCandidateSelection}

\begin{figure}[h]
    \centering
    \includegraphics[scale=0.7]{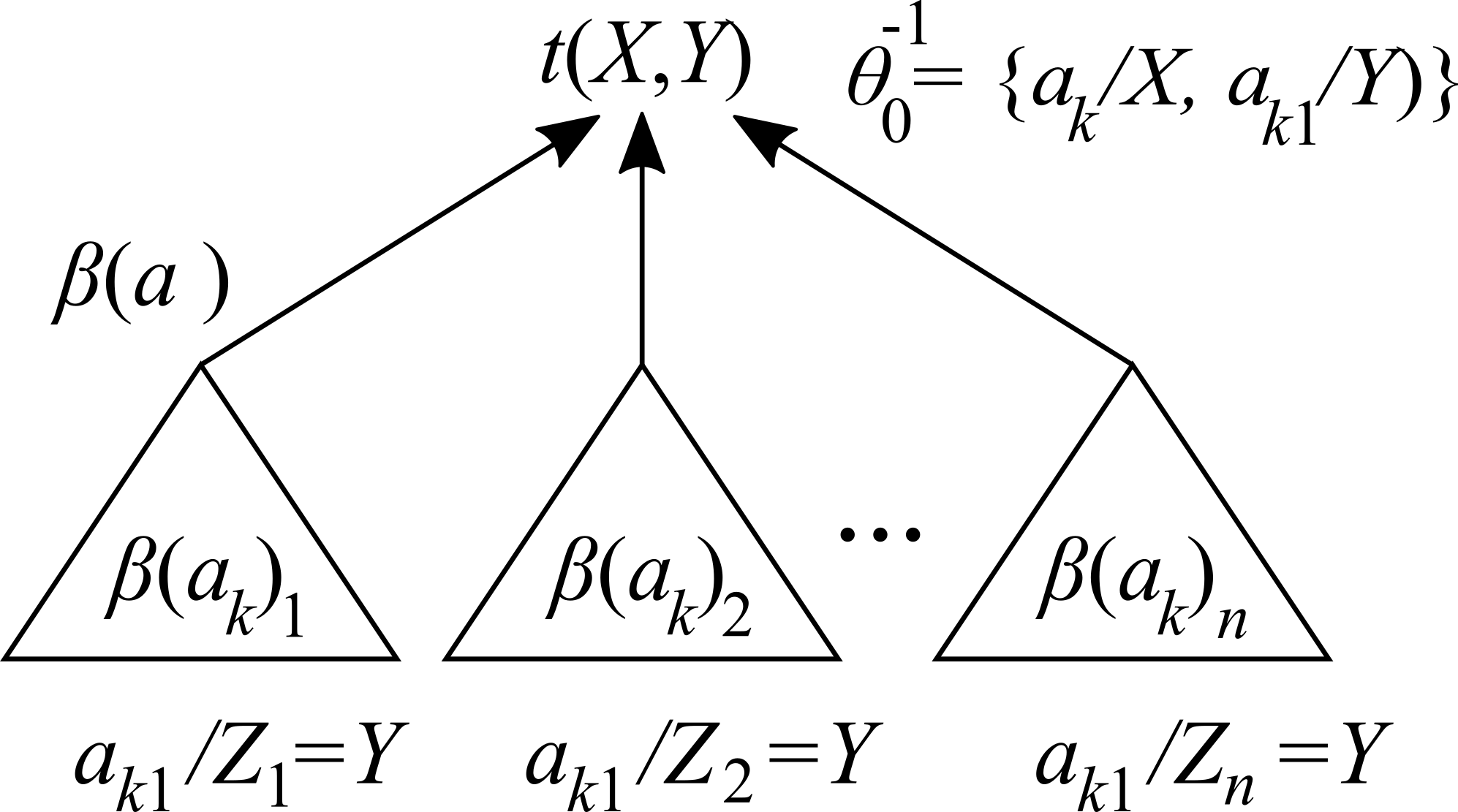}
    \caption{Space of hypotheses formed by Herbrand Base atoms.}
    \label{fig:HypoSpaceHB}
\end{figure}

Given the target $t(X,Y)$, $e^{+}$ : $t(a_k, a_{k1})$ and $e^{-}$ : $t(b_k, b_{k1})$. We access,
from the NeMuS of the BK, $\beta(a_k)$ and $\beta(a_{k1})$. The initial view of the space of possible
hypotheses that can be formed using atoms from the Herbrand Base and anti-unification is
illustrated in Figure~\ref{fig:HypoSpaceHB}. 

Each $\beta(a_{k})_i$ in a triangle represents a 
hypothesis formation branch that can be expanded following the bindings of the attribute in 
$e^{+}$. Some of them may allow the deduction of $e^{-}$, and thus \emph{inductive momentum} 
is applied to validate fetched atoms. 
 After adding an anti-unified literal from $\beta(a_k)_1$ into the premise of the hypothesis 
 being generated, say $H_1$, the next induction step will take a branch from the attribute-mates
of $a_k$ to compute $I_{\mu}$, generalize and so on. This is a depth-first walk across the
Herbrand Base. In the breadth-first walk the generation of  $H_1$ is postponed until all 
triangle branches have been initially exploited. For completeness sake it is implemented breadth-first.

\subsection{Computing $I_{\mu}$ and Linkages}
\label{sec:computeIm}

Accessing the bindings of constants is straightforward, we keep a loop selecting
the instances of the literals that appear until the last is verified. Basically it is running
while computing $I_{\mu}$ and moving across links of the sub-trees (triangles) from
 Figure~\ref{fig:HypoSpaceHB}.

If $I_{\mu}(q_1, \eta)^{a_k}_{b_k} = $ \emph{consistent} 
for all  $\eta \in \beta(b_k)$, then 
\begin{itemize}
 \item [] If $q_1(a_k,a_{k1}) \in \beta(a_k)$ and  $q_1(a_k,a_{k1}) \in \beta(a_{k1})$, then
         \subitem $H_1$: $t(X,Y) \leftarrow q_1 (X,Y)$, $\theta^{-1}_1 = \{a_k/X, a_{k1}/Y\}$
\item [] Else if $q_1(a_k,a_{k1}) \in \beta(a_k)$ and $q_1(a_k,a_{k1}) \not \in  \beta(a_k)$, then
          $H_1$: $t(X,Y) \leftarrow q_1 (X,Z_0)$,  $\theta^{-1}_1 = \{a_k/X, c/Z_0\}$
\item [] Otherwise, get another $q_j \in \beta(a_k)$ and repeat the process until there are no more
elements to test. In this case there is no hypothesis.
\end{itemize}                   
                   
\begin{figure}[h]
\centering
\includegraphics[scale=0.7]{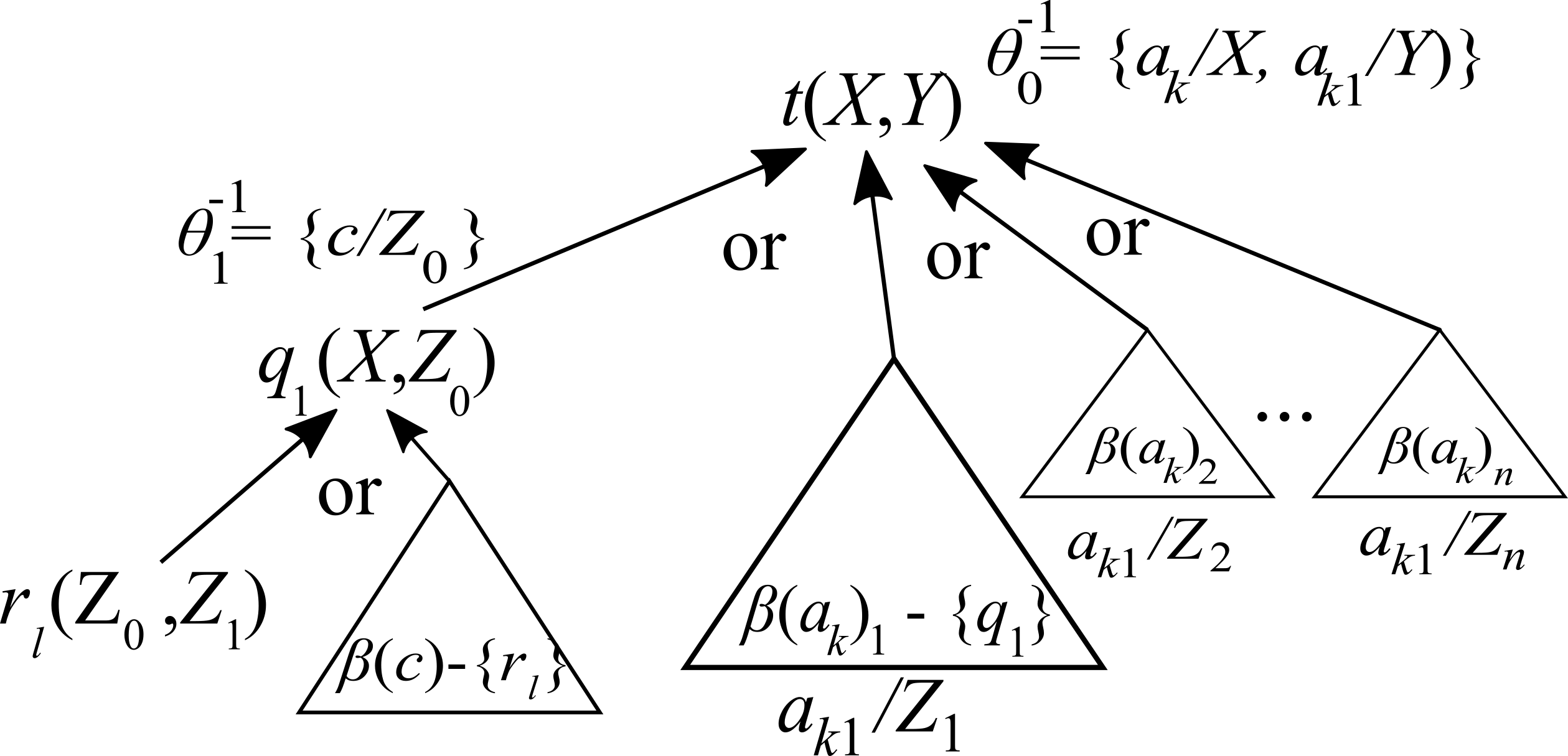}
\caption{Expanding space of hypotheses following $\beta(a_k)$ and  $\beta(c)$.}
\label{fig:HypoSpaceExpansion}
\end{figure}

For a consistent $H_1$, then there may exist $r_l \in \beta(c)$, and
\begin{itemize}
\item [a)] $r_l  \in \beta(a_{k1}):  r_l \not = q_1$,   
                $r_l(c, a_{k1})$ is an atom  from the Herbrand Base then for 
                $\theta^{-1}_1 =   \{a_k/X, c/Z_0, a_{k1}/Y\}$
                
               $H_1$: $t(X,Y) \leftarrow q_1 (X,Z_0) \land r_l(Z_0, Y)$.  (Chain in ILP) 
\item [b)] $r_l \not \in \beta(a_{k1})$:  path can only form 
                a \emph{long linear linkage} pattern. For non dyadic, if $I_{\mu}(r_l,\eta')$ is
                ok then expand hypotheses : $\beta(c) - \{r_l\}$. $H_1$'s body is added
                with $r_l(Z_0,Z_1)$ (see expansion illustrated 
                 in Figure~\ref{fig:HypoSpaceExpansion}).
\end{itemize}
                   
\subsection{``Bias" as Invention of Predicates}
\label{sec:biasAsInvention}

Amao performs a similarity training on NeMuS's weights using the vector
representation for each constant as well as for literals. Those with similar 
linkages end up with similar weight values associated to the argument 
they have and their position within them. Besides, bias may be used to add non 
targeted new predicates. 

\subsubsection*{Non user bias: ``automated" invention}

For this, it is necessary ``to invent" a predicate, say $p_0$,
such that $H_1$ becomes a closed hypothesis. For the sake of space $\theta^{-1}$
will be suppressed when anti-substituions are clear.
                 
\begin{itemize}
\item [] $H_1$: $t(X,Y) \leftarrow q_1(X,Z_0) \land p_{0}(Z_0,Y)$, 
             The invented predicate becomes
             the head of "invented hypothesis", as
                 
\item []$H_2$: $p_{0}(X,Y) \leftarrow r_l(X,Z_0)$ with $\theta^{-1}_1 = \{c/X, c_k/Z_0\}$,  
           and it becomes the \emph{current open hypothesis}. The search now is guided by 
           $\beta(c_k)$.
\end{itemize}

\subsubsection*{User defined bias for invention}

When an assumption that $r_l$ defines another concept, say $p_b$, then 
$H_1$'s body would have $p_b$ and $H_2$'s head would have $p_b$, rather then $p_0$. 
This would be something like {\tt assuming} $r_l(X,Y)$ {\tt defines}  $p_b(X,Y)$, then

\begin{itemize}	         
\item []$H_1$: $t(X,Y) \leftarrow q_1(X,Z_0) \land p_{b}(Z_0,Y)$, 
\item []$H_2$: $p_{b}(X,Y) \leftarrow r_l(X,Z_0)$
 \end{itemize}    
             
\begin{itemize}
\item Assuming $r_l = q_1$, i.e. both are the same predicate (concept region).
        \begin{enumerate}
        \item If $q_1(c,a_{k1}) \in BK$, and no bias given.
        
                Simple \emph{linear linkage} pattern (chain)
                \subitem $H_1$: $t(X,Y) \leftarrow q_1(X,Z_0) \land q_1(Z_0,Y)$,
                
                ``Shallow" \emph{recursive linkage} pattern (recursive tail)
                \subitem $H_1$: $t(X,Y) \leftarrow q_1(X,Z_0) \land t(Z_0,Y)$
                \subitem $H_2$: $t(X,Y) \leftarrow q_1(X,Y)$
                
                The order they are introduced into the set of clauses is unimportant
        \item If $q_1(c,a_{k1})  \not \in BK$ 
        \begin{enumerate}
        \item For bias and non dyadic theory: long linear linkage pattern
                 of the same concept would generate
                 
                 $H_1$: $t(X,Y) \leftarrow q_1(X,Z_0) \land q_1(Z_0,Z_1) \land \ldots \land q_1(Z_n,Y)$.

                 Instead, if $q_1(a_{k},c)$ and $q_1(c,a_{k1})$ region's weights are similar,
                 then invent of a recursive hypothesis.
                \subitem $H_1$: $t(X,Y) \leftarrow q_1(X,Z_0) \land t(Z_0,Y)$, 
                \subitem $H_2$: $t(X,Y) \leftarrow q_1(X,Y)$, 
                             
        \end{enumerate}        
        \end{enumerate}
\end{itemize}               


 As there can be many bindings, we close an open hypothesis for each possible
 combination of bindings.  Then, we keep computing the momentum and expanding 
 a new branch for each combination (as explained  in sections~\ref{sec:hypoCandidateSelection}
to~\ref{sec:biasAsInvention}). 

\subsection{A Running Example: the Family Tree}

{\bf Example 2}.
Consider the Family Tree, from \cite{muggleton15meta}. We may request to Amao the following

\begin{figure}[h!]
    \centering
    \includegraphics[scale=0.73]{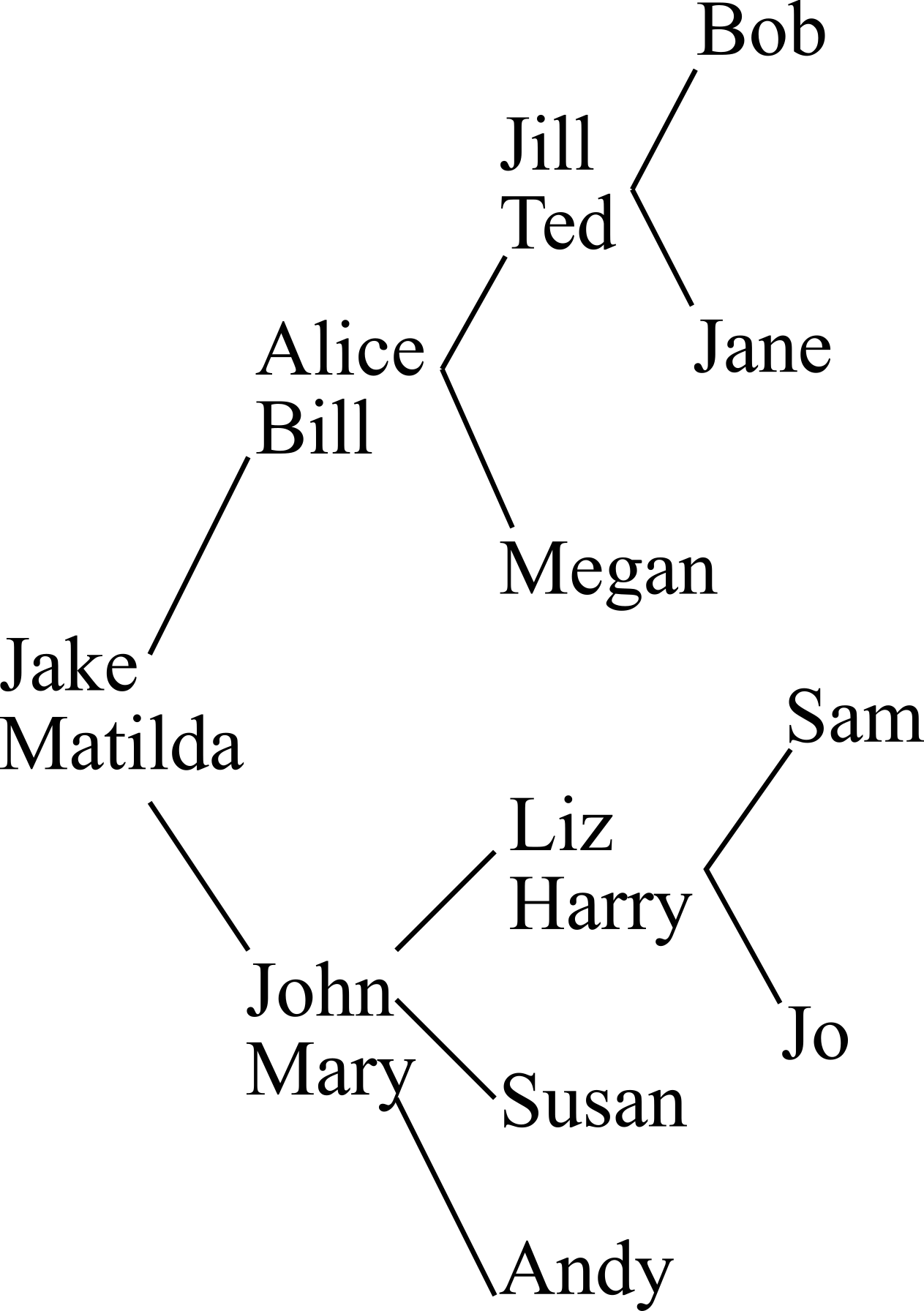}
    \caption{A simple family tree.}
    \label{fig:familyTreeMGTON}
\end{figure}


\noindent
\textbf{consider induction on} ancestor(X,Y) 
\textbf{knowing} ancestor(jake,bob)  
\textbf{assuming} father(X,Y) \textbf{or} mother(X,Y) 
\textbf{defines} parent(X,Y).

\begin{table}[th]
\centering
\begin{tabular}{cll}
\toprule
$i$ & atom/hypothesis               & $ \theta^{-1}_i$  / $\beta(a)$          \\
\midrule
0         & $ancestor(jake,bob) $  &  $\{jake/X, bob/Y\} $     \\
$H_0$ & $ancestor(X,Y) \leftarrow$     & \\
1         & $\beta_{1}(jake)$         & $father(jake,alice)$ \\   
           & $\beta_{1}(bob)$          &  $father(ted,bob)$   \\
           &  $I_{\mu} = consistent$  &  no hook \\
\hdashline
 bias  &    $father(X,Y)$             &   match both $\beta_{1}$ \\
  for     &  $parent(X,Y) $           &   rename variable \\
\hdashline
$H^{i}_0$         & $parent(X,Y) \leftarrow $         & $father(X,Y)$ \\   
           & $\{jake/X, bob/Y, alice/Z_0\} $          &  \\
  $H_1$ & $ancestor(X,Y) \leftarrow$     & $parent(X,Z_0)$ \\
                  &  $\beta_{1}(alice)$  &  $mother(alice, ted)$ \\

\hdashline
 bias  &    $mother(X,Y)$           &   match both $\beta_{1}(alice)$ \\
  for     &  $parent(X,Y) $           &   rename variable \\
\hdashline
  $H^{i}_1$  & $parent(X,Y) \leftarrow $   &  $mother(X,Y)$   \\
  $H_1$  & reaches maximun body size, & do not add another \\
              & $parent(Z_0, Z_1)$. Check for & region similarity \\
\hdashline
  $H^{i}_2$  & $ancestor(X,Y) \leftarrow $   &  $parent(X,Y)$   \\
 $H_1$ & $ancestor(X,Y) \leftarrow$     & $parent(X,Z_0)$ $\land$ \\
            &                                                & $ancestor(Z_0,Y)$ \\
\bottomrule
\end{tabular}
\end{table}


%% file: related.tex
\section{Related Work}\label{related}
Recent advances in Inductive Logic Programming (ILP) ease predicate invention by constraining  logical learning operations with higher-order meta-rules, expressions that describe the formats of the rules. These rules have order constraints associated to them (to ensure termination of the proof) and are provided to the meta-interpreter, which attempts to prove the examples. When successful at this task, it then saves the substitutions for existentially quantified variables in the meta-rules \cite{muggletonachievements}. This technique has been used to build Metagol \cite{muggleton14meta,muggleton15meta}, which has been successful in various examples. However, this approach tends to increase the generation of \textit{meaningless hypotheses} and, consequently, leads to a large hypotheses space. \cite{cropperijcai16} tackles this challenge by extending Metagol to support abstractions and invention, but it remains a problem. Amao takes a totally different approach by using NeMuS to perform Inductive Clause Learning (ICL). This work 
extends ICL by using the results of exploring weights of logical components, 
already present in NeMuS, to support inductive learning by expanding clause candidates with 
literals which passed in the inductive momentum check. This allows an efficient invention  
of predicates, including the learning of recursive hypotheses, while restricting the 
shape of the hypothesis by adding bias definitions or idiosyncrasies of the language.

%% file: conclusions.tex
\section{Discussion and Future Work}\label{conclusions}

This paper has shown how the Amao Shared NeMuS data structure can be used in predicate invention without the need to generate meaningless hypotheses. This is achieved via Inductive Clause Learning, with automatic predicate generation that takes advantage of the degree of importance of constant objects. As atomic object, constants of the Herbrand base had never called much attention for logical inference, but only to validate resolution through unification. Here, we showed how they can be used to guide the search for consistent hypothesis in two ways. 

First, by walking across their bindings from positive examples which are not rejected by inductive momentum with bindings of negative examples. Second, we use, as a heuristic to a faster generation of potentially recursive hypotheses, the maps or regions of similarities (item 2 of bias for invention), that constant bindings allow us to compute. Such maps as inductive mechanism is demonstrated in \cite{barretoSOIReasoningNeMuS19}. 

Future works will focus on making more efficient use of weighted structures of concepts and their composition to allow learning and reasoning of complex formulae, as well as dealing with noise, uncertainty, and possible worlds. We then aim to incorporate deep learning-like mechanisms by taking advantage of the inherently interconnected \textit{compound spaces} as a sort of layers for convolution when dealing with massive datasets.

\subsection*{Acknowledgement}